\pdfoutput=1
\documentclass{bmvc2k}

\usepackage{booktabs}
\usepackage{threeparttable}

\title{Expression Empowered ResiDen Network for Facial Action Unit Detection}

\addauthor{Shreyank Jyoti}{shreyank.jyoti@iitrpr.ac.in}{1}
\addauthor{Abhinav Dhall}{abhinav@iitrpr.ac.in}{1}
\addinstitution{Learning Affect and Semantic Image analysIs (LASII) Group,\\
Indian Institute of Technology Ropar\\
Punjab, India
}
    \runninghead{ResiDen}{Residue flow in Densenet}

\begin{document}

\maketitle

\begin{abstract}
The paper explores the topic of Facial Action Unit (FAU) detection in the wild. In particular, we are interested in answering the following questions: (1) how useful are residual connections across dense blocks for face analysis? (2) how useful is the information from a network trained for categorical Facial Expression Recognition (FER) for the task of FAU detection? The proposed network (ResiDen) exploits dense blocks along with residual connections and uses auxiliary information from a FER network. The experiments are performed on the EmotionNet and DISFA datasets. The experiments show the usefulness of facial expression information for AU detection. The proposed network achieves state-of-art results on the two databases. Analysis of the results for cross database protocol shows the effectiveness of the network.

\end{abstract}

%-------------------------------------------------------------------------
\section{Introduction}
\label{sec:intro}
Facial analysis task has recently become an area of the great interest for many researchers and industrial works. It automatically infers the information regarding emotional states, levels of engagement, pain detection, action unit detection, facial landmark detection and facial expression recognition from nonverbal behavior \cite{baltruvsaitis2016openface}. These activities show the depth in this area. With the interest in improving Human-Computer Interaction (HCI), the study of human behavior and its state of mind is a must \cite{fragopanagos2005emotion}. Expressions can be said as a gateway to humans state of mind for that particular instance. Thus for HCI, human's face plays a vital role. 

This paper aims to improve the facial action unit detection by drawing a relative bridge between expressions and the action units. Facial action units refer to the movement of the facial muscles such as raising eyebrows, opening mouth, lifting up cheeks etc.\ These movements can express the human state of mind or some actions such as eating, speaking, or communicating through facial gestures \cite{tian2001recognizing}. Thus, Action Units (AUs) convey a lot of information. AUs and facial muscle movements share different types of relationships: one to one, one to many, many to one \cite{zhang2013multi}. Not only some particular facial muscle movements represent some AUs, but several facial muscles movements can represent a particular AU and vice versa as well. These are set according to Facial Action Coding System (FACS) \cite{ekman1997face,facs}. FACS is an anatomically-based comprehensive system to annotate the visible facial muscle movements. 

Human minds are trained to observe the facial muscle movements and then to relate with the expressions \cite{facs}. Many expression recognition tasks have shown the superior performance with the help of AUs as the input features \cite{fasel2003automatic,pantic2004facial,li2013simultaneous,zhong2012learning,liu2013aware}. Here, we are interested to predict AUs with the help of categorical emotion classes based features.

Our motivation is to observe that whether expression based information could be beneficial for AUs prediction?
To this end, we performed our experiments on the DISFA \cite{mavadati2013DISFA} and the EmotioNet \cite{emotionet} dataset. We train a network that can deal with the dense blocks with residual sharing, for the task of FAU detection. Secondly, to utilize facial expression features, we train our proposed network on RAF-DB dataset for basic emotion recognition. Using this model, we extract second last layer features and tried to fuse these features in the model for action unit detection. Results show the superior performance of the proposed network over previous state-of-the-art results. Later on, we also observed the cross modality performance of the proposed network. 
The key contributions of this paper are:\begin{itemize}
  \item We propose a network ResiDen for action unit detection, that captures the significance of the dense blocks with residual sharing.
  \item We utilize facial expression features for facial action unit detection. Moreover an increase in accuracy is observed after merging two emotion categories (anger \& disgust). 
\end{itemize}

\section{Literature Review}

FAU detection is the task to observe the facial muscle movements. Many researchers have found interest towards the task. We are going to discuss some of the interesting and relevant works done in this area.

Previously, the detection of AU was done by following two main approaches \cite{lucey2006aam}. The first approach is based on obtaining facial landmarks or other information through pre-trained face model. This approach is called model-driven methods. Pre-trained networks like Active Appearance Models (AAMs) are trained to localize the facial landmarks \cite{lucey2007investigating}. These landmarks are found to be helpful for performing several tasks on the face image. These model suffers from person dependency and are not very accurate to person independent problems. This is a research topic in itself. The other approach is based on processing the entire facial image without collecting any prior analyses on the faces. These are called data-driven methods. Data driven methods can deal with person independent problems but a large number of training samples are required to deal with some challenges like illumination, head pose variation etc.

The AU prediction being an active area of research, still lacks in the intensity level predictions of AUs. To the best of our knowledge, DISFA \cite{mavadati2013DISFA} is one of the first dataset that has labelled AU intensity levels. Most of the dataset related to AUs are binary classification, which encounters the presence or the absence of a particular AU. Another direction is working on large-scale data, to this end the EmotioNet database \cite{emotionet} has been recently proposed.

Some previous works attempted to utilize the importance of both ResNet \cite{he2016deep} and Dense-\\Net \cite{huang2017densely}. Cheung et al. \cite {cheungdenresnet} proposed a network DenResNet where the last layer features of residual network and DenseNet are concatenated and are further used for the classification task on ILSVRC ImageNet dataset. The research was able to achieve state-of-the-art results. Proposed approach differs totally as we tried to share the residue between the blocks and utilize the importance of residual network in different manner.  

In 2017, Tong et al. \cite{tong2017image} proposed a network for Image super resolution using dense blocks. The approach involves a skip connection that forwards the information of initial convolutional layers and each dense block to the main latent space which furthers helps to generate a image with super resolution.  
In a recent Arxiv paper \cite{zhang2018residual}, a similar approach to ResiDen is proposed as Residual dense block (RDB). Tong et al. \cite{zhang2018residual} have implemented a hybrid network using RDN for producing a super-resolution image from low-resolution image.  

\section{Methodology}
We propose a network for dealing with the facial appearance-based attributes to predict the presence of Facial Action Units. Our network comprises of facial images and the emotion attributes to improve the AU detection accuracy. 

\subsection{Proposed Network}
Recently, it has been found that CNN perform well for the task AU detection \cite{li2017action}. To extend the network in depth, spatial information loss in CNN must be taken into consideration. It is observed that for face related vision problems, appearance based features are richer as compared to any other handcrafted features \cite{sariyanidi2015automatic}. Further, it is also possible that due to head movement in some cases, the facial point localization process can be noisy, which can be dealt with data augmentation. Now we discuss the structure of our proposed network. We create two channels (networks) for analyzing the AU: ResiDen Network and the Expression Network. The idea behind using these is to relate the information collected from the emotions and the facial features. We refer to the networks as ResiDen and Expression Net respectively.

 \begin{figure*}[t]
      \centering
        \includegraphics[scale=0.27]{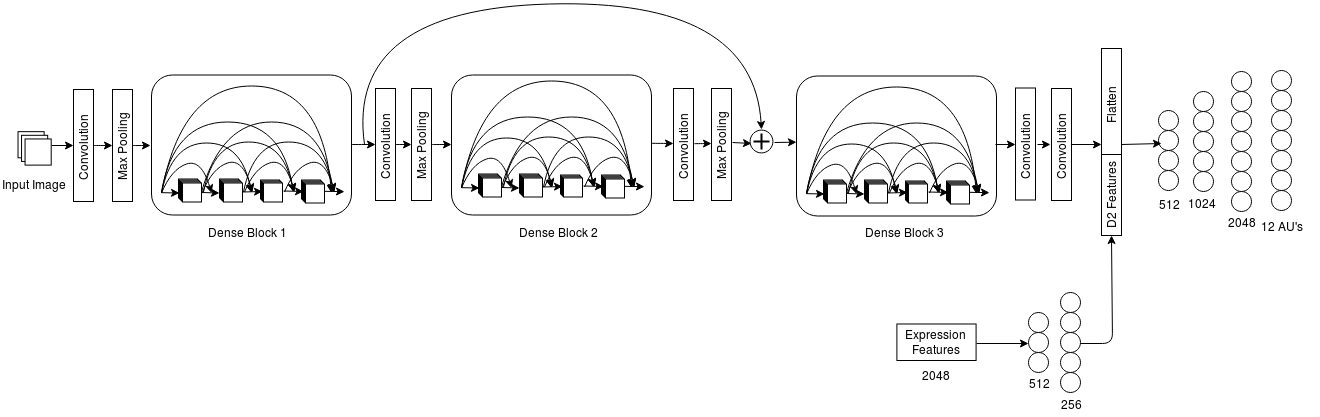}
      \vspace{0.9mm}
      \caption{The figure shows the pipeline for Action Unit Prediction. D2 refers to the Expression features after passing through two dense layers. }
      \label{fig:Network}
   \end{figure*}

\subsubsection{\textbf{ResiDen Network}}
The higher level features when observed seems to lose the relative spatial information, however, they are found to be the discriminative features for any task. On the other hand, lower level features are found to have the relative spatial information, however, they also contain noise, which can degrade the prediction quality. We attempt to create dense blocks with successive blocks holding prior blocks features. It is further observed that after a certain number as we increase the number of blocks, the network seems to fail. The reason behind the network failure is vanishing gradients. To control the gradient flow, we attempt to add few skip connections between the blocks. The gradients of consecutive blocks are summed up and then forwarded to the successive blocks.

\subsubsection{\textbf{Expression Net}} Expression network utilizes the importance of emotion features in context to Action Unit detection. Expression features are referred as the second last layer features of the network trained for the task of basic emotion recognition. The network, to compute the best expression features is finalized after several experiments. We train a basic CNN model with 4 convolution layers and 3 dense layers and compared the performance of expression recognition with the ResiDen model on RAF-DB dataset. The comparison is highlighted in Section \ref{experiment}. Once, we obtain the best emotion features we pass them through 2 dense layers and then concatenated with the facial features obtained after applying flatten layer to the ResiDen model. These features show a significant increase in the accuracy for the task of Facial Action Unit detection. The detailed architecture design of both the networks is discussed in Section \ref{experiment}.
%\subsection{Fine-tuning}

%\subsection{Noisy Labelling}

\subsection{Summing Up}
To integrate the models and to utilize the benefits of both the models we concatenate the features of both the  networks. The ResiDen returns 4096 features after flatten which are concatenated with the 256 emotion features after passing through two dense layers. These features are then finally used to predict the Action Units by applying 3 fully connected layers.
The training of the model is done by applying joint fine tuning with an aspect to improve the network weights as a whole rather than improving the weights of each individual network.

\section{Experiments}
\label{experiment}

The experiment section analyses the performance of the proposed network in comparison to some previous state-of-the-art results on DISFA and EmotioNet Dataset.

\subsection{Datasets Used}

\textbf{DISFA:} The Denver Intensity of Spontaneous Facial Action (DISFA) \cite{mavadati2013DISFA} dataset contains videos of 27 subjects (12 females and 15 males) of different age groups. The facial action unit is annotated at frame level by 2 FACS experts which justifies the significance of the dataset.  Each action unit is marked in the range of 0-5 according to the intensity of each AU. 0 marks the absence of the particular AU and 5 marks the highest intensity of the AU. 12 Action Units were marked for each frame.  Each video is of 4 minute duration and contains 4845 frames of $1024\times 768$ resolution.
The light illumination varies in each video.

% It is difficult to collect data with proper AU annotation as it must be annotated  only by the experts in FACS. This makes the data, one of the best datasets available for AU detection task.
% The video was recorded using PtGrey stereo imaging system.

\textbf{EmotioNet:} To utilize the  deep learning to its best, the large dataset is must. As per our knowledge, EmotioNet \cite{emotionet} is known to be the largest dataset for Action Unit detection. It contains nearly 9,00,000 images collected from wordnet using several words related to emotions. These images are automatically annotated for expression and AU labels. 11 AUs are annotated for each image.

% The dataset is distributed in train, validation and test set. Train dataset contains nearly 9,00,000 images whereas validation dataset contains 25,000 images. Test dataset contains 22,000 images which are further augmented to obtain 88,000 images. Results are discussed on validation dataset as test set is not available for performing experiments.

\textbf{RAF-DB:} 
Real-world Affective Faces Database (RAF-DB) \cite{li2017reliable} contains about 30,000 facial images downloaded from the Internet. 40 annotators contribute for the crowd source annotation. The images in the dataset are diversified in term of subject's age, gender, race, ethnicity, head poses, lighting conditions, occlusions, (e.g. glasses, facial hair or self-occlusion), post-processing operations (e.g. various filters and special effects), etc.
The RAF-DB is well known for its two subsets: basic emotions and compound emotions. As we need to train a model to extract the differentiability between the emotions, we chose to work on basic emotions only. There are 7 basic emotion classes for the RAF-DB.

% The dataset also provided the 42 facial landmarks out of which 5 were annotated accurately and the rest 37 were annotated automatically using trained AAM which is diversified among race, age range and gender attributes annotations per image.
%  Compound emotion refers to the multi label class like surprisingly happy.

\subsection {Data Pre-processing}

\textbf{EmotioNet:}
Based on the URLs shared by the dataset creators, we were able to download 6,00,000 approx images for the \emph{Train} set and 22,000 images for \emph{Validation} set. To extract the face location, we used the Multi-task Cascaded Convolutional Networks (MTCNN)  \cite{7553523} library. Each cropped face is then resized to $128 \times 128$ pixels and passed as an input to the network. 
\textbf{DISFA: } To process the video, 4845 frames are extracted from each videos. Using facial landmarks, we cropped the entire face by adding some threshold pixel value to the eyebrow landmarks as the forehead information is important for detection of some of the AUs. Each cropped face is then resized to $128 \times 128$ pixels and passed as an input to the network. For the \textbf{RAF-DB}, the pre-processing is similar to the DISFA dataset.

%  The ReLu can be mathematically expressed as :
%  \begin{equation}
% f(x)=max(0,x)
% \end{equation}
% The mathematical expression for swish is :
% \begin{equation}
% f(x)= x. sigmoid(x)
% \end{equation}

\subsection{ResiDen Architecture}
Now we discuss the details of the ResiDen Network. The network takes an image of size $128 \times 128$ as an input. Convolution with filter size  $3 \times 3$ and 48 kernels is applied to the input image. Swish activation function is used for all the networks. Batch Normalization is applied at each transition layer. The output of the convolution layer is then forwarded to max pooling layer with kernel size $2 \times 2$. After this, 3 dense blocks are applied with layers 12, 12, 36 respectively. The growth rate is set to 32. Transition layer after each dense block  is kept similar. $1 \times 1$ convolution is applied and later on it is passed through max pooling layer before entering the next dense block. The output of the second block is added up with the output of the first block similar to the skip connection approach and then passed as an input to the third dense block. The output from the last dense block is forwarded to the convolution layers with 128, 256 kernels, respectively. To prevent the over-fitting over convolutions we applied both L1 and L2 regularizer norm of 0.001 in both these convolution layers. The architecture of the network can be seen in fig \ref{fig:Network}.          
  To check the performance of ResiDen alone, the output of the last convolution is flattened and fully connected layer with 512, 1024 and 2048 nodes is applied. These 2048 features are used for the final prediction of the task. Dropout of 0.4 and 0.2 is applied on the last two dense layers respectively. This ensures to get rid of the overfitting over the dense features.

\subsection{Expression Network}
The ResiDen with same architecture is trained on RAF-DB dataset for the emotion recognition task. Once the model is trained, we use it to extract emotion features of the DISFA and EmotioNet dataset. The expression features are extracted from the second last layer of the network which computes 2048 features. This feature vector endeavors to encode the different facial changes for different emotions. To pass these features to the final proposed model, we pass it through dense layers with nodes 512 and 256 respectively, in order to compute the best 256 emotion features. 

These 256 emotion features are concatenated with the features obtained after applying the flatten layer to the last convolution layer of the ResiDen network. These features are used for final AU detection task by applying 3 fully connected layer with 512, 2048 and 2048 nodes respectively. Further down, we have discussed the performance achieved by applying two different networks (ResiDen and CNN) as feature extractors.

\textbf{Impact of Merging Classes:}
It is noted from different facial expressions that the geometric structure of the face and the facial muscle movements has large difference among some facial expressions such as happy and sad. However, the same is not true for classes such as anger and disgust \cite{li2017reliable}. Facial expression recognition fails mainly for the class group of anger \& disgust and sad \& disgust as observed in the confusion matrix given by \cite{li2017reliable}. Based on this observation, we hypothise that a classifier should be able to distinguish between a fused expression category (eg.\ anger \& disgust) and other expressions. The reason to merge disgust \& anger is that they may have similar intensities for some facial action units \cite{li2017reliable}. It is observed in the experiment section that this improves the overall performance of the action unit detection. The performance with expression features when computed with 6 and 7 classes are compared in Section \ref{h}.

\textbf{Activation Function:} All the experiments are performed using Swish \cite{ramachandran2017swish} as an activation function. Swish shows an improvement over many experiments in the domain of image classification or deep learning approaches. To justify its performance, Ramachandran
et al. \cite{ramachandran2017swish} stated that ReLU being the most commonly used activation lacks as it allows the gradient flow for only the positive values and ignores the negative values totally. Swish in contrast is a smooth function which deals with all the values to predict the output. Swish is bounded from below but unbounded at above similar to ReLU but is non-monotonic in nature. It behaves as ReLU for the values near to extreme but it provides a linearity for the values close to average.

\subsection{Results}\label{h}

\subsubsection{Performance Units} 
\space

\textbf{DISFA:}
All the experiments on DISFA \cite{mavadati2013DISFA} are compared on accuracy basis. Accuracy can measure the count of true positives and false positives over the entire dataset. It ensure the prediction of the presence of the the correct AUs.
The accuracy for each AU can be mathematically calculated by 
\begin{equation}
accuracy_{i}=\frac{(true\  positives_{i} + true \  negatives_{i})}  {total\ samples}
\end{equation}

Here, i refers to the particular AU class. True positives are the presence of AUs predicted correctly. 
True negatives are the absence of AUs predicted correctly. To obtain the final accuracy for the AU prediction the mean is calculated for all the classes.

\textbf{EmotioNet:}
% Precision is computed as True Positive/(True Positive + False Positive).
% Recall is computed as True Positive/(True Positive + False Negative).
%  F1 score is calculated by    
% \begin{equation}
% F_{\beta_{i}}=(1+ \beta)^{2} \frac{precision_{i} + recall_{i}}  {\beta^{2} precision_{i} + recall_{i}}
% \end{equation}
% Here, $\beta$ is the constant value and i refers to a particular AU class. 
All the experiments on EmotioNet \cite{emotionet} are compared on the basis of final score. Final score is the average of accuracy and the $F_{1}$ score. Final score not only ensure the correct prediction of the AUs but also utilizes the importance of precision and recall. Precision ensures the correct detected samples over predicted samples.  Recall ensures the the correct detected samples over true samples. Precision and Recall jointly computes F1 score. The final score for each AU class is computed as 
\begin{equation}
final\  score_{i}=\frac{(accuracy_{i}+ F_{1_{i}})} {2}
\end{equation}

The final score for all the classes of AU prediction is the mean of all final score values.

All the experiments on RAF-DB \cite{li2017reliable} are compared on the basis of accuracy. Since the emotion recognition is a single label class, the accuracy is measured by $\frac{Correct\  Predicted\  Images}{Total\  Images}$.

\subsubsection{Observations}

We compared the performance of the ResiDen with Residue networks and the Dense networks. The results can be seen in Table \ref{t2}. We tried different architectures for all the networks and observed a significant growth on applying ResiDen. This ensures the novelty over previously available networks. Here we can notice that the Resnet was able to achieve the max performance of 70.92 \% accuracy on DISFA and 0.6711 final score value for EmotioNet dataset. Densenet performs better by achieving max of 73.68 \% accuracy on DISFA and 0.6913 final score for EmotioNet dataset. ResiDen performs the best by achieving 76.74 \% accuracy on DISFA and 0.7110 final score for EmotioNet dataset. It is observed that the ResiDen with 3 blocks obtains the best results and thus all the other experiments are performed by setting the number of blocks to 3. To obtain the best emotion features we compared the results on RAF-DB. The results can be seen in Table \ref{t1}. CNN refers to a basic convolution network. 

% \usepackage{multirow}
% \usepackage[normalem]{ulem}

\begin{table}[tbh]
\begin{threeparttable}
\centering
\begin{minipage}{.49\linewidth}
\centering
\small{
%\caption{(a)}
%\caption{Resuts on DISFA dataset. The performance is noted in terms of accuracy.  $EF_{RD/n}$ & $EF_{CNN/n}$ refers to the emotion features extracted using ResiDen and CNN trained on n classes respectively.}
%\label{t1}
\begin{tabular}{|l|c|}
\hline
%{\textit{\textbf{Network}}}                & { \textit{\textbf{DISFA \\ (accuracy in \%)}}} \\ 
%T & G \\ \hline

{ \textit{\textbf{Network}}}                & { \textit{\textbf{DISFA}}} \\ 
                                               & { \textit{\textbf{(accuracy in \%)}}} \\ \hline

SVM\cite{mavadati2013DISFA}                                     & 74.90                                          \\ \hline
EAC-Net \cite{li2017eac}                                       & 80.60                                          \\ \hline
$C_{1024}$\cite{ghasemi2017deep}                                     & 74                                          \\ \hline
$\phi(C 14)$\cite{ghasemi2017deep}                                       & 76                                          \\ \hline
Resnet 34                                      & 70.21                                          \\ \hline
Resnet 50                                      & 70.80                                          \\ \hline
Resnet 101                                     & 70.92                                          \\ \hline
Densenet (3 blocks)                            & 73.11                                          \\ \hline
Densenet (5 blocks)                            & 73.68                                          \\ \hline
Densenet (7 blocks)                            & 73.07                                          \\ \hline
ResiDen (3 blocks)                             & 76.74                                          \\ \hline
ResiDen (5 blocks)                             & 75.69                                          \\ \hline
ResiDen (7 blocks)                             & 75.33                                          \\ \hline
ResiDen (Data Aug)                             & 78.32                                          \\ \hline
Residen +$EF_{RD/7}$                           & 79.04                                          \\ \hline
\textbf{ResiDen + $EF_{RD/6}$}                          & \textbf{79.87}                                          \\ \hline
ResiDen + $EF_{CNN/7}$                         & 78.83                                          \\ \hline
ResiDen + $EF_{CNN/6}$                         & 79.41                                          \\ \hline
Cross Dataset                                  & 68.26                                          \\ \hline
\end{tabular}

\begin{tablenotes}
      \small
      \item (a) Resuts on DISFA dataset.
    \end{tablenotes}

}
\end{minipage} 
\hspace{0.6cm}
\begin{minipage}{.49\linewidth}
\centering
\small{
%\caption{(a)}
%\caption{Results on EmotioNet dataset. The performance is noted in terms of final score. EF refers to the emotion features. }
%\label{t3}
\begin{tabular}{|l|c|}
\hline
{ \textit{\textbf{Network}}}                & { \textit{\textbf{EmotioNet}}} \\ 
                                               & { \textit{\textbf{(final score)}}} \\ \hline
I2R-CCNU-NTU-2 \cite{emotionet}                                & 0.72                                           \\ \hline
JHU \cite{emotionet}                                           & 0.71                                            \\ \hline
I2R-CCNU-NTU-1 \cite{emotionet}                                 & 0.70                                           \\ \hline
I2R-CCNU-NTU-3 \cite{emotionet}                                & 0.69                                           \\ \hline
Resnet 34                                      & 0.66                                          \\ \hline
Resnet 50                                      & 0.67                                          \\ \hline
Resnet 101                                     & 0.67                                          \\ \hline
Densenet (3 blocks)                            & 0.68                                          \\ \hline
Densenet (5 blocks)                            & 0.68                                          \\ \hline
Densenet (7 blocks)                            & 0.69                                          \\ \hline
ResiDen (3 blocks)                             & 0.71                                          \\ \hline
ResiDen (5 blocks)                             & 0.71                                          \\ \hline
ResiDen (7 blocks)                             & 0.71                                          \\ \hline
ResiDen (Data Aug)                             & 0.72                                            \\ \hline
Residen + $EF_{RD/7}$                          & 0.73                                            \\ \hline
\textbf{ResiDen + $EF_{RD/6}$}                          & \textbf{0.74}                                            \\ \hline
ResiDen + $EF_{CNN/7}$                         & 0.73                                            \\ \hline
ResiDen + $EF_{CNN/6}$                         & 0.73                                            \\ \hline
Cross Dataset                                  & 0.64                                            \\ \hline
\end{tabular}
\begin{tablenotes}
      \small
      \item (a) Resuts on EmotioNet dataset.
    \end{tablenotes}
}
\end{minipage} 
\vspace{3mm}
\caption{ Performance of our proposed network on different datasets. Here, $EF_{RD/n}$ \& $EF_{CNN/n}$ refers to the emotion features extracted using ResiDen and CNN trained on n classes respectively.}\label{t2}
\end{threeparttable}
\end{table}

% \usepackage[normalem]{ulem}

\textbf{CNN Architecture:}
The network consists of four convolution and three dense layers. The convolution with filters 48, 128, 256, 256 is applied to the successive convolution layers respectively. The output of first, third, and fourth convolution is pooled using max as the pooling function with kernel size $2 \times 2$. The output of the last layers is flattened and is used to predict the final emotion. Three fully connected layers are applied as an intermediate to flatten and the final prediction. The fully connected layer contains 512, 512, 2048 hidden units respectively. Dropout  of 40 \% and 20 \% is applied after first and second fully connected layer.

\begin{table}[tbh]
\centering

\begin{tabular}{|l|c|}

\hline
\multicolumn{1}{|c|}{\textit{\textbf{Network}}} & \multicolumn{1}{c|}{\textit{\textbf{Accuracy (\%)}}} \\ \hline
M-SVM                                           & 65.12                                           \\ \hline
DLP-CNN                                         & 74.22                                           \\ \hline
ResiDen (7 classes)                             & 76.54                                                \\ \hline
CNN (7 classes)                                 &  78.23                                               \\ \hline
ResiDen (6 classes)                             &    83.83                                             \\ \hline
CNN (6 classes)                                 &  85.77                                               \\ \hline
\end{tabular}
\vspace{3mm}
\caption{Performance of the Expression Network on the RAF-DB dataset.}\label{t1}
\end{table}

The CNN achieves the highest accuracy on RAF-DB. The reason could be the dataset doesn't contains information to pass through the deep network. The deep network could face the issue of vanishing gradients after a certain layers. We tried our best to tackle the vanishing gradients issue but still it has some threshold. The last layers must be generating some noise which leads to the failure in the best learned parameters. Since the ResiDen and CNN both were able to beat the state-of-the-art results for RAF-DB, we use features extracted from both the networks and compared the final results.
The Table \ref{t1} also shows the results on the merged class. By 6 classes, we mean that the emotion `angry' and `disgust' is treated as a single emotion.

% \usepackage{multirow}
% \usepackage[normalem]{ulem}

The final results of our proposed network is seen in Table \ref{t2}. The results are obtained on DISFA and EmotioNet. The data augmentation helped in achieving better results on both the datasets. For facial dataset, scaling is one of the most important data augmentation technique as it focuses the face images more closely and highlights the facial features \cite{sariyanidi2015automatic}. Rotation of faces helps in handling the faces with roll movements \cite{trad2012facial}. Augmentation includes rotation of images with random angles -15$^\circ$ to 0$^\circ$ \& 0$^\circ$ to 15$^\circ$ and scaling by a scaling factor of 0.1. As we notice an increase in performance after applying data augmentation, all the other results mentioned for ResiDen are with data augmentation. A significant increase in performance can be observed when the emotion features comes into play. The results also ensures the effectiveness of merging the classes for expression. A surprising performance can be noticed with the expression features extracted using ResiDen. CNN performing individually well on RAF-DB shows less improvement to the results as compared to the expression features extracted using ResiDen. The reason for the contradictory performance of the network could be either the CNN undergoes overfitting or there is a data variation in validation set.

The trained model shows its effectiveness by providing good results on the cross dataset experiment. The cross dataset refers to the training done on a particular dataset and testing is applied on some other dataset. Here the trained model with DISFA when used to test the results on EmotioNet it achieves a final score of 0.6409. The trained model with EmotioNet achieves the 68.26 \% accuracy on DISFA dataset. As the model trained for DISFA dataset predicts the 12 AUs, we neglected the prediction of the AU corresponding to the "Lip corner depressor" as all the other AU classes were same for both the datasets. Similar approach is applied for the case of model trained on EmotioNet. We really can't compare the results directly but atleast have some idea of the performance of the network when applied to any real world problem. These experiments validate the novelty of the ResiDen network individually. Further, the results also show the usefulness of expression features in increasing the performance of AU detection task.

\begin{figure*}[t]
      \centering
        \includegraphics[scale=0.27]{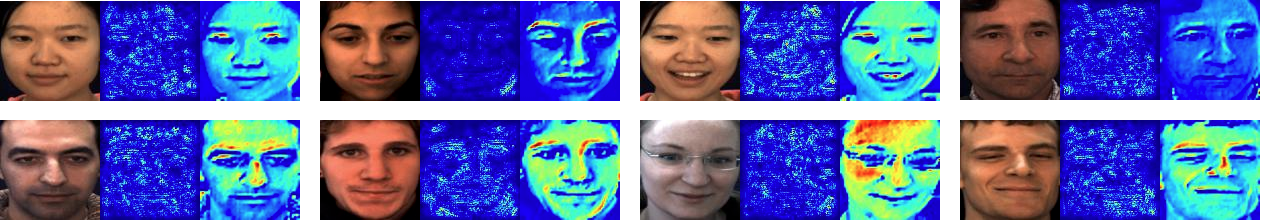}
      \vspace{0.9mm}
      \caption{Visualization of facial emotion. Each set of three images shows the original image, saliency map and class activation map respectively. }
      \label{vis_face}
   \end{figure*}

 \textbf{Visualization:}
Our best model for DISFA dataset is used to visualize the learning of the network. Class activation map as shown in Figure \ref{vis_face} depicts the predicted AU. The confidence of the prediction is max on the red marks.  
%%%%%%%%%%%%%%%%%%%%%%%%%%%%%%%%%%%%%%%%%%%%%%%%%%%%%%%%%%%%%%%%%%%%%%%%%%%%%%%%%%%%%%%%%%%%%%%%%%%%%%%%%%%
\section{Conclusions}
The intensive experiments shows the importance of expression features for the prediction of Action Units. To add on, a hybrid network sharing the residue between the blocks could be found beneficial. The systems shows the improvement in the state-of-the-art for Action Unit prediction. The task of Action Unit annotation is difficult and expensive as it must be annotated only by the experts in FACS. An automatic system can ease the effort and the cost for the annotation task. For such systems, expression features can be one of the most important cues.

% \section{Future Works }
\section{Acknowledgement}
  We would like to thanks NVIDIA for donating Titan X for our research work. 
  We would also like to thank Dr. Mahoor and Dr. Martinez for providing us with the dataset and there valuable response.
\bibliography{egbib}
\end{document}